\documentclass[10pt,twocolumn,letterpaper]{article}

\usepackage{cvpr}
\usepackage{times}
\usepackage{epsfig}
\usepackage{graphicx}
\usepackage{amsmath}
\usepackage{amssymb}
\usepackage{soul}

\usepackage[backend=biber,
            hyperref=true,
            url=false,
            isbn=false,
            doi=false,
            backref=false,
            style=ieee,
            natbib=true,
            citestyle=numeric-comp,
            sorting=nyt,
            maxcitenames=1,
            mincitenames=1,
            block=none]{biblatex}

\usepackage[breaklinks=true,letterpaper=true,colorlinks,bookmarks=false]{hyperref}

\cvprfinalcopy

\begin{document}

\title{DenseFusion: 6D Object Pose Estimation by Iterative Dense Fusion}

\author{Chen Wang$^{2}$ \qquad Danfei Xu$^{1}$ \qquad Yuke Zhu$^{1}$ \qquad Roberto Mart\'{i}n-Mart\'{i}n$^{1}$\\Cewu Lu$^{2}$ \qquad Li Fei-Fei$^{1}$ \qquad Silvio Savarese$^{1}$\\
{\tt\small $^{1}$Department of Computer Science, Stanford University}\\
{\tt\small $^{2}$Department of Computer Science, Shanghai Jiao Tong University}
}

\maketitle

\begin{abstract}
A key technical challenge in performing 6D object pose estimation from RGB-D image is to fully leverage the two complementary data sources. Prior works either extract information from the RGB image and depth separately or use costly post-processing steps, limiting their performances in highly cluttered scenes and real-time applications. In this work, we present DenseFusion, a generic framework for estimating 6D pose of a set of known objects from RGB-D images. DenseFusion is a heterogeneous architecture that processes the two data sources individually and uses a novel dense fusion network to extract pixel-wise dense feature embedding, from which the pose is estimated. Furthermore, we integrate an end-to-end iterative pose refinement procedure that further improves the pose estimation while achieving near real-time inference. Our experiments show that our method outperforms state-of-the-art approaches in two datasets, YCB-Video and LineMOD. We also deploy our proposed method to a real robot to grasp and manipulate objects based on the estimated pose. Our code and video are available at https://sites.google.com/view/densefusion/.
\end{abstract}

\section{Introduction}

6D object pose estimation is the crux to many important real-world applications, such as robotic grasping and manipulation~\cite{collet2011moped,zhu2014single,tremblay2018deep}, autonomous navigation~\cite{kitti,xu2017pointfusion,mv3d}, and augmented reality~\cite{tangle,marchand2016pose}. Ideally, a solution should deal with objects of varying shape and texture, show robustness towards heavy occlusion, sensor noise, and changing lighting conditions, while achieving the speed requirement of real-time tasks. The advent of cheap RGB-D sensors has enabled methods that infer poses of low-textured objects even in poorly-lighted environments more accurately than RGB-only methods. Nonetheless, it is difficult for existing methods to satisfy the requirements of accurate pose estimation and fast inference simultaneously.

Classical approaches first extract features from RGB-D data and perform correspondence grouping and hypothesis verification~\cite{hinterstoisser2012model,Hinterstoier2011MultimodalTF,rios2013discriminatively,kehl2016deep,tejani2014latent,wohlhart2015learning,brachmann2014learning}. However, the reliance on handcrafted features and fixed matching procedures have limited their empirical performances in presence of heavy occlusion and lighting variation. Recent success in visual recognition has inspired a family of data-driven methods that use deep networks for pose estimation from RGB-D inputs, such as PoseCNN~\cite{xiang2017posecnn} and MCN~\cite{li2018unified}.

\begin{figure}[t]
	\centering
	\includegraphics[width=.98\linewidth]{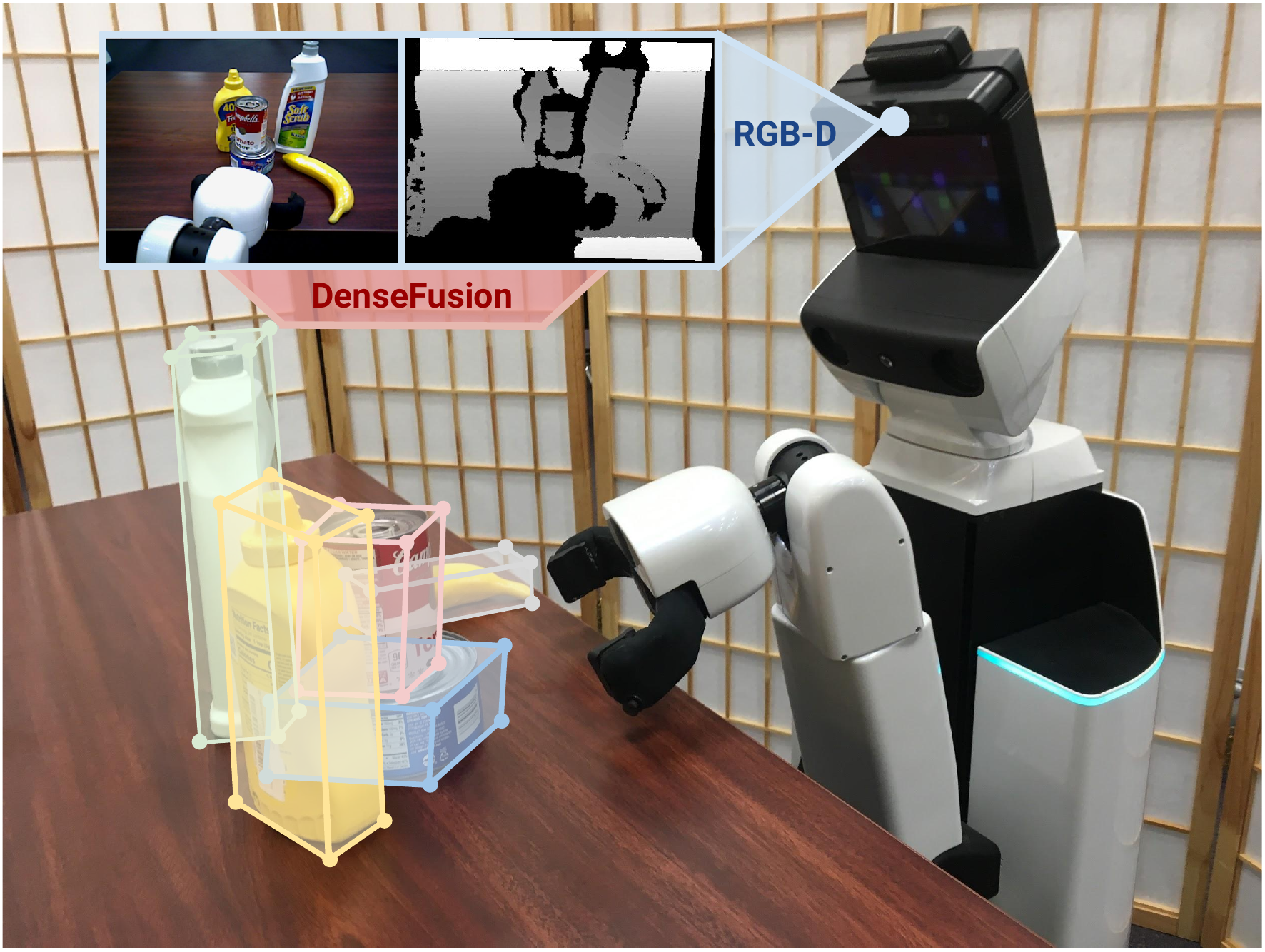}
	\caption{We develop an end-to-end deep network model for 6D pose estimation from RGB-D data, which performs fast and accurate predictions for real-time applications such as robot grasping and manipulation.}
	\label{fig:pull}
	\vspace{-15pt}
\end{figure}

However, these methods require elaborate post-hoc refinement steps to fully utilize the 3D information, such as a highly customized Iterative Closest Point (ICP)~\cite{Besl1992AMF} procedure in PoseCNN and a multi-view hypothesis verification scheme in MCN. These refinement steps cannot be optimized jointly with the final objective and are prohibitively slow for real-time applications. In the context of autonomous driving, a third family of solutions has been proposed to better exploit the complementary nature of color and depth information from RGB-D data with end-to-end deep models, such as Frustrum PointNet~\cite{qi2017frustum} and PointFusion~\cite{xu2017pointfusion}. These models have achieved good performances in driving scenes and the capacity of real-time inference. However, as we demonstrate empirically, these methods fall short under heavy occlusion, which is common in manipulation domains.

In this work, we propose an end-to-end deep learning approach for estimating 6-DoF poses of known objects from RGB-D inputs. 
The core of our approach is to embed and fuse RGB values and point clouds at per-pixel level, as opposed to prior work which uses image crops to compute global features~\cite{xu2017pointfusion} or 2D bounding boxes~\cite{qi2017frustum}. This per-pixel fusion scheme enables our model to explicitly reason about the local appearance and geometry information, which is essential to handle heavy occlusion. Furthermore, we propose an iterative method which performs pose refinement within the end-to-end learning framework. This greatly enhances model performance while keeping the inference speed real-time.

We evaluate our method in two popular benchmarks for 6D pose estimation, YCB-Video~\cite{xiang2017posecnn} and LineMOD~\cite{Hinterstoier2011MultimodalTF}.
We show that our method outperforms the state-of-the-art PoseCNN after ICP refinement~\cite{xiang2017posecnn} by 3.5\% in pose accuracy while being 200x faster in inference time. In particular, we demonstrate its robustness in highly cluttered scenes thanks to our novel dense fusion method. Last, we also showcase its utility in a real robot task, where the robot estimates the poses of objects and grasp them to clear up a table.

In summary, the contributions of this work are two-fold: First, we present a principled way to combine color and depth information from the RGB-D input. We augment the information of each 3D point with 2D information from an embedding space learned for the task and use this new color-depth space to estimate the 6D pose. Second, we integrate an iterative refinement procedure within the neural network architecture, removing the dependency of previous methods of a post-processing ICP step.

\section{Related Work}
\noindent\textbf{Pose from RGB images.} Classical methods rely on detecting and matching keypoints with known object models~\cite{aubry2014seeing, collet2011moped,zhu2014single,rothganger20063d,ferrari2006simultaneous}. Newer methods address the challenge by learning to predict the 2D keypoints~\cite{pavlakos20176,suwajanakorn2018discovery,tremblay2018deep,tekin18,brachmann2014learning} and solve the poses by P\emph{n}P~\cite{fischler1981random}. Though prevail in speed-demanding tasks, these methods become unreliable given low-texture or low-resolution inputs. Other methods propose to directly estimate objects pose from images using CNN-based architectures~\cite{tulsiani2015viewpoints,schwarz2015rgb}. Many such methods focus on orientation estimation: \citet{subcnn,xiang2015data} learns a viewpoint-aware pose estimator by clustering 3D features from object models. \citet{mousavian20163d} predicts 3D object parameters and recovers poses by single-view geometry constraints. \citet{sundermeyer2018implicit} implicitly encode orientation in a latent space and in test time find the best match in a codebook as the orientation prediction. However, pose estimation in 3D remains a challenge for the lack of depth information. Our method leverages both image and 3D data to estimate object poses in 3D in an end-to-end architecture.

\vspace{1mm}
\noindent
\textbf{Pose from depth / point cloud.}
Recent studies have proposed to directly tackle the 3D object detection problem in discretized 3D voxel spaces. For example, \citet{song2014sliding,song2016deep} generate 3D bounding box proposals and estimate the poses by featuring the voxelized input with 3D ConvNets. Although the voxel representation effectively encodes geometric information, these methods are often prohibitively expensive:~\cite{song2016deep} takes nearly 20 seconds for each frame.

More recent 3D deep learning architectures have enabled methods that directly performs 6D pose estimation on 3D point cloud data. As an example, both Frustrum PointNets~\cite{qi2017frustum} and VoxelNet~\cite{zhou2017voxelnet} use a PointNet-like~\cite{qi2016pointnet} structure and achieved state-of-the-art performances on the KITTI benchmark~\cite{kitti}. Our method also makes use of similar architecture. However, unlike urban driving applications for which point cloud alone provides enough information, generic object pose estimation tasks such as the YCB-Video dataset~\cite{xiang2017posecnn} demands reasoning over both geometric and appearance information. We address such a challenge by proposing a novel 2D-3D sensor fusion architecture.

\vspace{1mm}
\noindent
\textbf{Pose from RGB-D data.} 
Classical approaches extract 3D features from the input RGB-D data and perform correspondence grouping and hypothesis verification~\cite{hinterstoisser2012model,Hinterstoier2011MultimodalTF,rios2013discriminatively,kehl2016deep,tejani2014latent,wohlhart2015learning,brachmann2014learning}. However, these features are either hard-coded~\cite{hinterstoisser2012model,Hinterstoier2011MultimodalTF,rios2013discriminatively} or learned by optimizing surrogate objectives~\cite{tejani2014latent,wohlhart2015learning,brachmann2014learning} such as reconstruction~\cite{kehl2016deep} instead of the true objective of 6D pose estimation. Newer methods such as PoseCNN~\cite{xiang2017posecnn} directly estimates 6D poses from image data.
\citet{li2018unified} further fuses the depth input as an additional channel to a CNN-based architecture. However, these approaches rely on expensive post-processing steps to make full use of 3D input. In comparison, our method fuses 3D data to 2D appearance feature while retaining the geometric structure of the input space, and we show that it outperforms~\cite{xiang2017posecnn} on the YCB-Video dataset~\cite{xiang2017posecnn} without the post-processing step. 

Our method is most related to PointFusion~\cite{xu2017pointfusion}, in which geometric and appearance information are fused in a heterogeneous architecture. We show that our novel local feature fusion scheme significantly outperforms PointFusion's naive fusion-by-concatenation method. In addition, we use a novel iterative refinement method to further improve the pose estimation.

\section{Model}
\begin{figure*}[ht]
	\centering
	\includegraphics[width=0.8\linewidth]{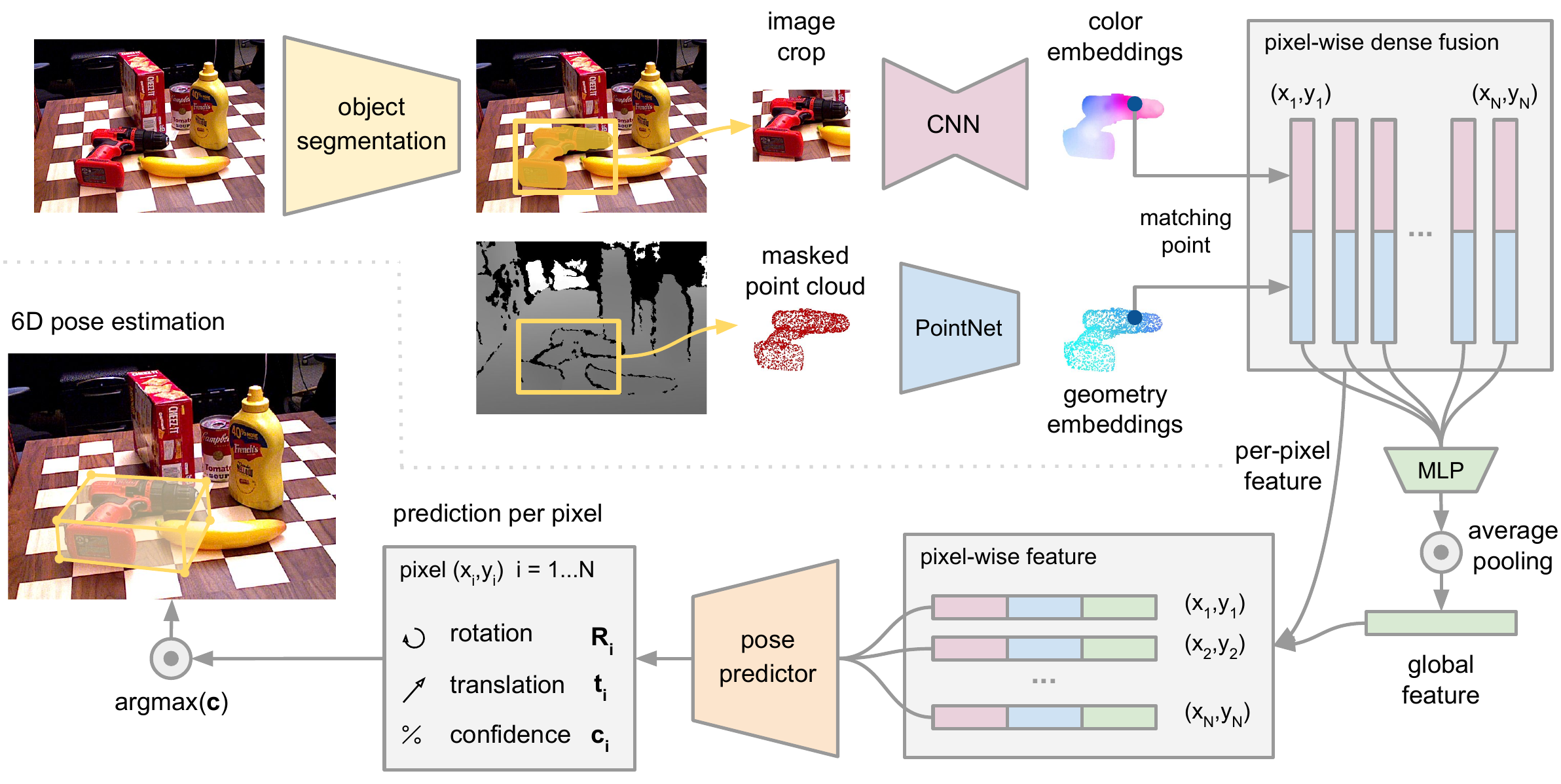}
	\caption{\textbf{Overview of our 6D pose estimation model.} Our model generates object segmentation masks and bounding boxes from RGB images. The RGB colors and point cloud from the depth map are encoded into embeddings and fused at each corresponding pixel. The pose predictor produces a pose estimate for each pixel and the predictions are voted to generate the final 6D pose prediction of the object. (The iterative procedure of our approach is not depicted here for simplicity)}
	\label{fig:overall}
\end{figure*}

Our goal is to estimate the 6D pose of a set of known objects present in an RGB-D image of a cluttered scene.
Without loss of generality, we represent 6D poses as homogeneous transformation matrix, $p \in SE(3)$. 
In other words, a 6D pose is composed by a rotation $R \in SO(3)$ and a translation $t \in \mathbb{R}^3$, $p=[R|t]$.
Since we estimate the 6D pose of the objects from camera images, the poses are defined with respect to the camera coordinate frame.

Estimating the pose of a known object in adversarial conditions (e.g. heavy occlusion, poor lighting, \ldots) is only possible by combining the information contained in the color and depth image channels. 
However, the two data sources reside in different spaces. Extracting features from heterogeneous data sources and fusing them appropriately is the key technical challenge in this domain. 

We address this challenge with (1) a heterogeneous architecture that processes color and depth information differently, retaining the native structure of each data source (Sec.~\ref{ssec::cpf}), and (2) a dense pixel-wise fusion network that performs color-depth fusion by exploiting the intrinsic mapping between the data sources (Sec.~\ref{ssec:pwdf}). Finally, the pose estimation is further refined with a differentiable iterative refinement module (Sec.~\ref{ssec::refine}). In contrast to the expensive post-hoc refinement steps used in~\cite{xiang2017posecnn,li2018unified}, our refinement module can be trained jointly with the main architecture and only takes a small fraction of the total inference time.

\subsection{Architecture Overview}
\label{sec::overview}

Fig.~\ref{fig:overall} illustrates the overall proposed architecture. The architecture contains two main stages. The first stage takes color image as input and performs semantic segmentation for each known object category. Then, for each segmented object, we feed the masked depth pixels (converted to 3D point cloud) as well as an image patch cropped by the bounding box of the mask to the second stage. 

The second stage processes the results of the segmentation and estimates the object's 6D pose. It comprises four components: a) a fully convolutional network that processes the color information and maps each pixel in the image crop to a color feature embedding, b) a PointNet-based~\cite{qi2016pointnet} network that processes each point in the masked 3D point cloud to a geometric feature embedding, c) a pixel-wise fusion network that combines both embeddings and outputs the estimation of the 6D pose of the object based on an unsupervised confidence scoring, and d) an iterative self-refinement methodology to train the network in a curriculum learning manner and refine the estimation result iteratively. Fig. \ref{fig:overall} depicts a), b) and c) and Fig. \ref{fig:refinement} illustrates d). The details our architecture are described below.

\subsection{Semantic Segmentation}
The first step is to segment the objects of interest in the image. Our semantic segmentation network is an encoder-decoder architecture that takes an image as input and generates an $N+1$-channelled semantic segmentation map. Each channel is a binary mask where active pixels depict objects of each of the $N$ possible known classes. The focus of this work is to develop a pose estimation algorithm. Thus we use an existing segmentation architecture proposed by~\cite{xiang2017posecnn}.

\subsection{Dense Feature Extraction}
\label{ssec::cpf}

The key technical challenge in this domain is the correct extraction of information from the color and depth channels and their synergistic fusion. Even though color and depth present a similar format in the RGB-D frame, their information resides in different spaces. Therefore, we process them separately to generate color and geometric features from embedding spaces that retain the intrinsic structure of the data sources.

\noindent\textbf{Dense 3D point cloud feature embedding:}
Previous approaches have used CNN to process the depth image as an additional image channel~\cite{li2018unified}. However, such method neglects the intrinsic 3D structure of the depth channel. Instead, we first convert the segmented depth pixels into a 3D point cloud using the known camera intrinsics, and then use a PointNet-like architecture to extract geometric features. 

PointNet by~\citet{qi2016pointnet} pioneered the use of a symmetric function (max-pooling) to achieve permutation invariance in processing unordered point sets. The original architecture takes as input a raw point cloud and learns to encode the information about the vicinity of each point and of the point cloud as a whole. The features are shown to be effective in shape classification and segmentation~\cite{qi2016pointnet} and pose estimation~\cite{xu2017pointfusion,qi2017frustum}. We propose a geometric embedding network that generates a dense per-point feature by mapping each of the $P$ segmented points to a $d_{geo}$-dimensional feature space. We implement a variant of PointNet architecture that uses average-pooling as opposed to the commonly used max-pooling as the symmetric reduction function.

\noindent\textbf{Dense color image feature embedding:}
The goal of the color embedding network is to extract per-pixel features such that we can form dense correspondences between 3D point features and image features. The reason for forming these dense correspondences will be clear in the next section. The image embedding network is a CNN-based encoder-decoder architecture that maps an image of size $H\times W\times 3$ into a $H\times W\times d_{rgb}$ embedding space. Each pixel of the embedding is a $d_{rgb}$-dimensional vector representing the appearance information of the input image at the corresponding location.

\subsection{Pixel-wise Dense Fusion}
\label{ssec:pwdf}

So far we have obtained dense features from both the image and the 3D point cloud inputs; now we need to fuse the information. A naive approach would be to generate a global feature from the dense color and depth features from the segmented area. However, due to heavy occlusion and segmentation errors, the set of features from previous step may contain features of points/pixels on other objects or parts of the background. Therefore, blindly fusing color and geometric features globally would degrade the performance of the estimation. In the following we describe a novel pixel-wise\footnote{Since the mapping between pixels and 3D points is unique, we will use interchangeably \emph{pixel-fusion} and \emph{point-fusion}.} dense fusion network that effectively combines the extracted features, especially for pose estimation under heavy occlusion and imperfect segmentation.

\noindent\textbf{Pixel-wise dense fusion:}
The key idea of our dense fusion network is to perform local per-pixel fusion instead of global fusion so that we can make predictions based on \emph{each} fused feature. In this way, we can potentially select the predictions based on the visible part of the object and minimize the effects of occlusion and segmentation noise. Concretely, our dense fusion procedure first associates the geometric feature of each point to its corresponding image feature pixel based on a projection onto the image plane using the known camera intrinsic parameters. The obtain pairs of features are then concatenated and fed to another network to generate a fixed-size global feature vector using a symmetric reduction function. While we refrained from using a single global feature for the estimation, here we enrich each dense pixel-feature with the global densely-fused feature to provide a global context.

We feed each of the resulting per-pixel features into a final network that predicts the object's 6D pose. In other words, we will train this network to predict one pose from each densely-fused feature. The result is a set of $P$ predicted poses, one per feature. This defines our first learning objective, as we will see in Sec.~\ref{ssec::loss}. We will now explain our approach to learn to choose the best prediction in a self-supervised manner, inspired by the work by \citet{xu2017pointfusion}.

\noindent\textbf{Per-pixel self-supervised confidence:}
We would like to train our pose estimation network to decide which pose estimation is likely to be the best hypothesis based on the specific context. To do so, we modify the network to output a confidence score $c_i$ for each prediction in addition to the pose estimation predictions. We will have to reflect this second learning objective in the overall learning objective, as we will see at the end of the next section.

\subsection{6D Object Pose Estimation}
\label{ssec::loss}
Having defined the overall network structure, we now take a closer look at the learning objective. We define the pose estimation loss as the distance between the points sampled on the objects model in ground truth pose and corresponding points on the same model transformed by the predicted pose.  Specifically, the loss to minimize for the prediction \textbf{per dense-pixel} is defined as
\begin{align}
L^{p}_{i} &= \frac{1}{M}\sum_{j}||(Rx_j + t) - (\hat{R_{i}}x_j + \hat{t_{i}})||
\end{align}
where $x_j$ denotes the $j^{th}$ point of the $M$ randomly selected 3D points from the object's 3D model, $p = [R|t]$ is the ground truth pose, and $\hat{p_{i}} = [\hat{R_{i}}|\hat{t_{i}}]$ is the predicted pose generated from the fused embedding of the $i^{th}$ dense-pixel. 

The above loss function is only well-defined for asymmetric objects, where the object shape and/or texture determines a unique canonical frame. Symmetric objects have more than one and possibly an infinite number of canonical frames, which leads to ambiguous learning objectives. Therefore, for symmetric objects, we instead minimize the distance between each point on the estimated model orientation and the \emph{closest} point on the ground truth model. The loss function becomes:
\begin{align}
L^{p}_{i} &= \frac{1}{M}\sum_{j}\min_{0 < k < M}||(Rx_j + t) - (\hat{R_{i}}x_k + \hat{t_{i}})||
\end{align}

Optimizing over all predicted per dense-pixel poses would be to minimize the sum of the per dense-pixels losses: $L = \frac{1}{N}\sum_{i}L^p_{i}$. However, as explained before, we would like our network to learn to balance the confidence among the per dense-pixel predictions. To do that we weight the per dense-pixel loss with the dense-pixel confidence, and add a second confidence regularization term:
\begin{align}
  \label{eq:confidence}
  L &= \frac{1}{N}\sum_{i}(L^p_{i} c_{i} - w \log(c_{i})),
\end{align}
where $N$ is the number of randomly sampled dense-pixel features from the $P$ elements of the segment and $w$ is a balancing hyperparameter. Intuitively, low confidence will result in low pose estimation loss but would incur high penalty from the second term, and vice versa. We use the pose estimation that has the highest confidence as the final output.

\subsection{Iterative Refinement}
\label{ssec::refine}

The iterative closest point algorithm (ICP)~\cite{Besl1992AMF} is a powerful refinement approach used by many 6D pose estimation methods~\cite{xiang2017posecnn,kehl2017ssd,sundermeyer2018implicit}. However, the best-performing ICP implementations are often not efficient enough for real-time applications. Here we propose a neural network-based iterative refinement module that can improve the final pose estimation result in a fast and robust manner.

\begin{figure}[t!]
	\centering
	\includegraphics[width=1.\linewidth]{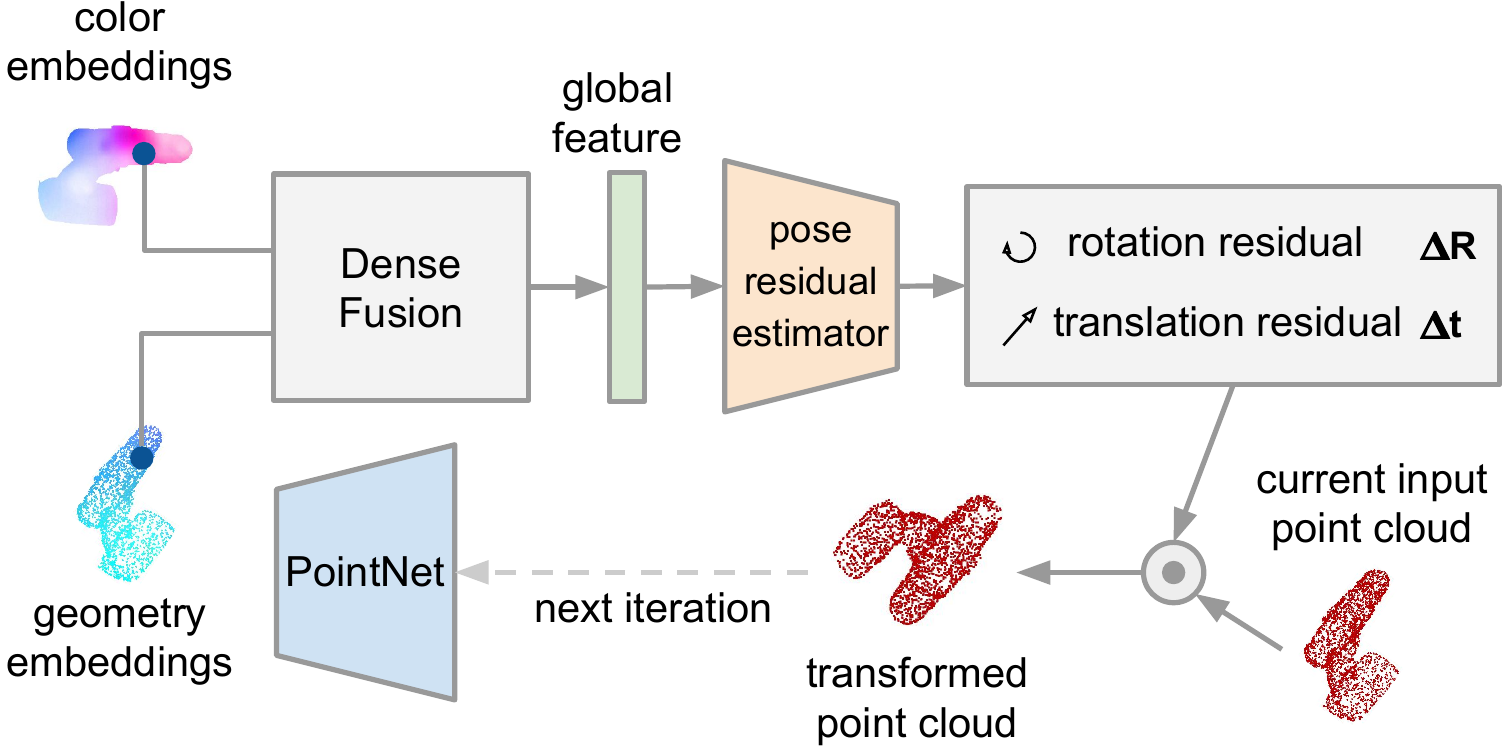}
	\caption{\textbf{Iterative Pose Refinement.} We introduce an network module that refines the pose estimation in an iterative procedure.}
	\label{fig:refinement}
\end{figure}

The goal is to enable the network to correct its own pose estimation error in an iterative manner. The challenge here is training the network to refine the previous prediction as opposed to making new predictions. To do so, we must include the prediction made in a previous iteration as part of the input to the next iteration. Our key idea is to consider the previously predicted pose as an estimate of canonical frame of the target object and transform the \textit{input} point cloud into this estimated canonical frame.  This way, the transformed point cloud implicitly encodes the estimated pose. We then feed the transformed point cloud back into the network and predict a \textit{residual pose} based on the previously estimated pose.  This procedure can be applied iteratively and generate potentially finer pose estimation each iteration. 

The procedure is illustrated in Fig.~\ref{fig:refinement}. Concretely, we train a dedicated pose residual estimator network to perform the refinement given the initial pose estimation from the main network. At each iteration, we reuse the image feature embedding from the main network and perform dense fusion with the geometric features computed for the new transformed point cloud. The pose residual estimator uses as input a global feature from the set of fused pixel features. After $K$ iterations, we obtain the final pose estimation as the concatenation of the per-iteration estimations:
\begin{align}
\hat{p} = [R_K|t_K] \cdot [R_{K - 1}|t_{K - 1}] \cdot \dots \cdot [R_{0}|t_{0}]
\end{align}

The pose residual estimator can be trained jointly with the main network. However, the pose estimation at the beginning of the training is too noisy for it to learn anything meaningful. Thus in practice, the joint training starts after the main network has converged.

\begin{table*}
\small
\centering
\caption{Quantitative evaluation of 6D pose (ADD-S\cite{xiang2017posecnn}) on YCB-Video Dataset. Objects with bold name are symmetric.}
\begin{tabular}{l|c|c|c|c|c|c|c|c|c|c}
\hline
                         & \multicolumn{2}{c}{PointFusion}~\cite{xu2017pointfusion} & \multicolumn{2}{c}{PoseCNN+ICP}~\cite{xiang2017posecnn} & \multicolumn{2}{c}{Ours (single)} & \multicolumn{2}{c}{Ours (per-pixel)} & \multicolumn{2}{c}{Ours (iterative)} \\ \hline
                         & AUC       & \textless{}2cm      & AUC            & \textless{}2cm & AUC        & \textless{}2cm       & AUC          & \textless{}2cm         & AUC              & \textless{}2cm    \\ \hline
002\_master\_chef\_can   & 90.9      & 99.8                & 95.8           & 100.0          & 93.9       & 100.0                & 95.2         & 100.0                  & \textbf{96.4}    & 100.0             \\
003\_cracker\_box        & 80.5      & 62.6                & 92.7           & 91.6           & 90.8       & 98.4                 & 92.5         & 99.3                   & \textbf{95.5}    & \textbf{99.5}     \\
004\_sugar\_box          & 90.4      & 95.4                & \textbf{98.2}  & 100.0          & 94.4       & 99.2                 & 95.1         & 100.0                  & 97.5             & 100.0             \\
005\_tomato\_soup\_can   & 91.9      & 96.9                & 94.5           & 96.9           & 92.9       & 96.7                 & 93.7         & 96.9                   & \textbf{94.6}    & 96.9              \\
006\_mustard\_bottle     & 88.5      & 84.0                & \textbf{98.6}  & 100.0          & 91.2       & 97.8                 & 95.9         & 100.0                  & 97.2             & 100.0             \\
007\_tuna\_fish\_can     & 93.8      & 99.8                & \textbf{97.1}  & 100.0          & 94.9       & 100.0                & 94.9         & 100.0                  & 96.6             & 100.0             \\
008\_pudding\_box        & 87.5      & 96.7                & \textbf{97.9}  & 100.0          & 88.3       & 97.2                 & 94.7         & 100.0                  & 96.5             & 100.0             \\
009\_gelatin\_box        & 95.0      & 100.0               & \textbf{98.8}  & 100.0          & 95.4       & 100.0                & 95.8         & 100.0                  & 98.1             & 100.0             \\
010\_potted\_meat\_can   & 86.4      & 88.5                & \textbf{92.7}  & \textbf{93.6}  & 87.3       & 91.4                 & 90.1         & 93.1                   & 91.3             & 93.1              \\
011\_banana              & 84.7      & 70.5                & \textbf{97.1}  & 99.7           & 84.6       & 62.0                 & 91.5         & 93.9                   & 96.6             & \textbf{100.0}    \\
019\_pitcher\_base       & 85.5      & 79.8                & \textbf{97.8}  & 100.0          & 86.9       & 80.9                 & 94.6         & 100.0                  & 97.1             & 100.0             \\
021\_bleach\_cleanser    & 81.0      & 65.0                & \textbf{96.9}  & 99.4           & 91.6       & 98.2                 & 94.3         & 99.8                   & 95.8             & \textbf{100.0}    \\
\textbf{024\_bowl}                & 75.7      & 24.1                & 81.0           & 54.9           & 83.4       & 55.4                 & 86.6         & 69.5                   & \textbf{88.2}    & \textbf{98.8}     \\
025\_mug                 & 94.2      & 99.8                & 95.0           & 99.8           & 90.3       & 94.7                 & 95.5         & \textbf{100.0}         & \textbf{97.1}    & \textbf{100.0}    \\
035\_power\_drill        & 71.5      & 22.8                & \textbf{98.2}  & \textbf{99.6}  & 83.1       & 64.2                 & 92.4         & 97.1                   & 96.0             & 98.7              \\
\textbf{036\_wood\_block}         & 68.1      & 18.2                & 87.6           & 80.2           & 81.7       & 76.0                 & 85.5         & 93.4                   & \textbf{89.7}    & \textbf{94.6}     \\
037\_scissors            & 76.7      & 35.9                & 91.7           & 95.6           & 83.6       & 75.1                 & 96.4         & \textbf{100.0}         & \textbf{95.2}    & \textbf{100.0}    \\
040\_large\_marker       & 87.9      & 80.4                & 97.2           & 99.7           & 91.2       & 88.6                 & 94.7         & 99.2                   & \textbf{97.5}    & \textbf{100.0}    \\
\textbf{051\_large\_clamp}        & 65.9      & 50.0                & \textbf{75.2}  & 74.9           & 70.5       & 77.1                 & 71.6         & 78.5                   & 72.9             & \textbf{79.2}     \\
\textbf{052\_extra\_large\_clamp} & 60.4      & 20.1                & 64.4           & 48.8           & 66.4       & 50.2                 & 69.0         & 69.5                   & \textbf{69.8}    & \textbf{76.3}     \\
\textbf{061\_foam\_brick}         & 91.8      & 100.0               & \textbf{97.2}  & 100.0          & 92.1       & 100.0                & 92.4         & 100.0                  & 92.5             & 100.0             \\ \hline
MEAN                      & 83.9      & 74.1                & 93.0           & 93.2           & 88.2       & 87.9                 & 91.2         & 95.3                   & \textbf{93.1}    & \textbf{96.8}  \\ \hline
\end{tabular}
\label{exp:ycb}
\end{table*}

\section{Experiments}
\label{sec:exp}

In the experimental section, we would like to answer the following questions: (1) How does the dense fusion network compare to naive global fusion-by-concatenation? (2) Is the dense fusion and prediction scheme robust to heavy occlusion and segmentation errors? (3) Does the iterative refinement module improve the final pose estimation? (4) Is our method robust and efficient enough for downstream tasks such as robotic grasping?

To answer the first three questions, we evaluate our method on two challenging 6D object pose estimation datasets: YCB-Video Dataset~\cite{xiang2017posecnn} and LineMOD~\cite{Hinterstoier2011MultimodalTF}. The YCB-Video Dataset features objects of varying shapes and texture levels under different occlusion conditions. Hence it's an ideal testbed for our occlusion-resilient multi-modal fusion method. The LineMOD dataset is a widely-used dataset that allows us to compare with a broader range of existing methods. We compare our method with state-of-the-art methods~\cite{sundermeyer2018implicit,kehl2017ssd} as well as model variants. To answer the last question, we deploy our model to a real robot platform and evaluate the performance of a robot grasping task that uses the predictions from our model.

\subsection{Datasets}
\label{ssec:dsm}

\noindent\textbf{YCB-Video Dataset.} The YCB-Video Dataset~\citet{xiang2017posecnn} features 21 YCB objects~\citet{alli2015TheYO} of varying shape and texture. The dataset contains 92 RGB-D videos, where each video shows a subset of the 21 objects in different indoor scenes. The videos are annotated with 6D poses and segmentation masks. We follow prior work~\cite{xiang2017posecnn} and split the dataset into 80 videos for training and 2,949 key frames chosen from the rest 12 videos for testing and include the same 80,000 synthetic images released by ~\cite{xiang2017posecnn} in our training set. In our experiments, we compare with the result of ~\cite{xiang2017posecnn} after depth refinement(ICP) and learning-based depth method ~\cite{xu2017pointfusion}.

\noindent\textbf{LineMOD Dataset.}
The LineMOD dataset ~\citet{Hinterstoier2011MultimodalTF} consists of 13 low-textured objects in 13 videos. It is widely adopted by both classical methods~\cite{vidal20186d, drost2010model, buch2017rotational} and recent learning-based approaches~\cite{sundermeyer2018implicit, tekin18, li2018deepim}. We use the same training and testing set as prior learning-based works~\cite{rad2017bb8, tekin18, li2018deepim} without additional synthetic data and compare with the best ICP-refined results of the state-of-the-art algorithms.

\begin{figure*}[ht]
	\centering
	\includegraphics[width=1.0\linewidth]{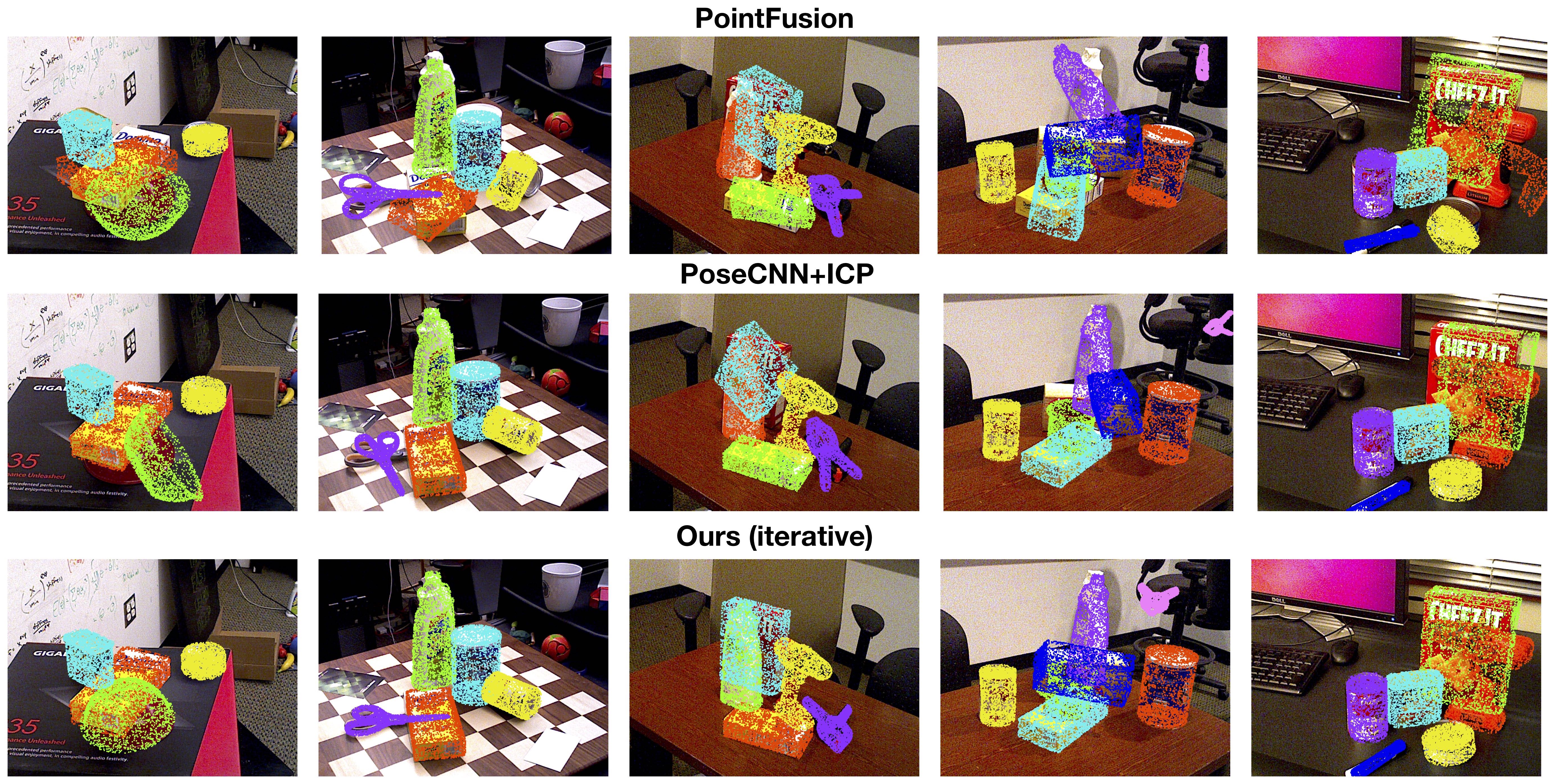}
	\caption{\textbf{Qualitative results on the YCB-Video Dataset.} All three methods shown here are tested with the same segmentation masks as in PoseCNN. Each object point cloud in different color are transformed with the predicted pose and then projected to the 2D image frame. The first two rows are former RGB-D methods and the last row is our approach with dense fusion and iterative refinement (2 iterations).}
	\label{exp:qualitative_ycb}
\end{figure*}

\subsection{Metrics}
\label{ssec:met}

We use two metrics to report on the YCB-Video Dataset. The average closest point distance (ADD-S)~\cite{xiang2017posecnn} is an ambiguity-invariant pose error metric which takes care of both symmetric and non-symmetric objects into an overall evaluation. Given the estimated pose $[\hat{R}|\hat{t}]$ and ground truth pose $[R|t]$, ADD-S calculates the mean distance from each 3D model point transformed by $[\hat{R}|\hat{t}]$ to its closest neighbour on the target model transformed by $[R|t]$. We report the area under the ADD-S curve (AUC) following PoseCNN~\cite{xiang2017posecnn}. We follow prior work and set the maximum threshold of AUC to be 0.1m. We also report the percentage of ADD-S smaller than 2cm (\textless{}2cm), which measures the predictions under the minimum tolerance for robot manipulation (2cm for most of the robot grippers).

For the LineMOD dataset, we use the Average Distance of Model Points (ADD)~\cite{hinterstoisser2012model} for non-symmetric objects and ADD-S for the two symmetric objects (\textit{eggbox} and \textit{glue}) following prior works~\cite{hinterstoisser2012model,sundermeyer2018implicit,tekin18}.

\subsection{Implementation Details}
\label{ssec::implementation}
The image embedding network consists of a Resnet-18 encoder followed by $4$ up-sampling layers as the decoder. The PointNet architecture is an MLP followed by an average-pooling reduction function. Both color and geometric dense feature embedding are of dimension $128$. We choose $w = 0.01$ for Eq.~\ref{eq:confidence} by empirical evaluation. The iterative pose refinement module consists of a $4$ fully connected layers that directly output the pose residual from the global dense feature. We use the $2$ refinement iterations for all experiments.

\subsection{Architectures}
We compare four model variants that showcase the effectiveness of our design choices.

\noindent~\textbullet~\textit{PointFusion}~\cite{xu2017pointfusion} uses a CNN to extract a fixed-size feature vector and fuse by directly concatenating the image feature with the geometry feature. The rest of the network is similar to our architecture. The comparison to this baseline demonstrates the effectiveness of our dense fusion network.

\noindent~\textbullet~\textit{Ours (single)} uses our dense fusion network, but instead of performing per-point prediction, it only outputs a single prediction using the global feature vector.

\noindent~\textbullet~\textit{Ours (per-pixel)} performs per-pixel prediction based on each densely fused feature.

\noindent~\textbullet~\textit{Ours (iterative)} is our complete model that uses the iterative refinement (Sec.~\ref{ssec::refine}) on top of \textit{Ours (per-pixel)}.

\subsection{Evaluation on YCB-Video Dataset}
\label{sec:eval_ycb}

Table~\ref{exp:ycb} shows the evaluation results for all the 21 objects in the YCB-Video Dataset. We report the \texttt{ADD-S AUC(<0.1m)} and the \texttt{ADD-S<2cm} metrics on PoseCNN~\cite{xiang2017posecnn} and our four model variants.  To ensure a fair comparison, all methods use the same segmentation masks as in PoseCNN~\cite{xiang2017posecnn}. Among our model variants, \textit{Ours (Iterative)} achieves the best performance. Our method is able to outperform PoseCNN + ICP\cite{xiang2017posecnn} even without iterative refinement. In particular, \textit{Ours (Iterative)} outperforms PoseCNN + ICP by $3.5\%$ on the \texttt{ADD-S<2cm} metric.

\noindent\textbf{Effect of dense fusion} Both of our dense fusion baselines (\textit{Ours (single)} and \textit{Ours (per-pixel)}) outperform \textit{PointFusion} by a large margin, which shows that dense fusion has a clear advantage over the global fusion-by-concatenation method used in \textit{PointFusion}. 

\noindent\textbf{Effect of iterative refinement} Table~\ref{exp:ycb} shows that our iterative refinement improves the overall pose estimation performance. In particular, it significantly improves the performances for texture-less symmetric object, e.g., \texttt{bowl} (29\%), \texttt{banana} (6\%), and \texttt{extra\_large\_clamp} (6\%) which suffer from orientation ambiguity.

\begin{figure}[t]
	\centering
	\includegraphics[width=.95\linewidth]{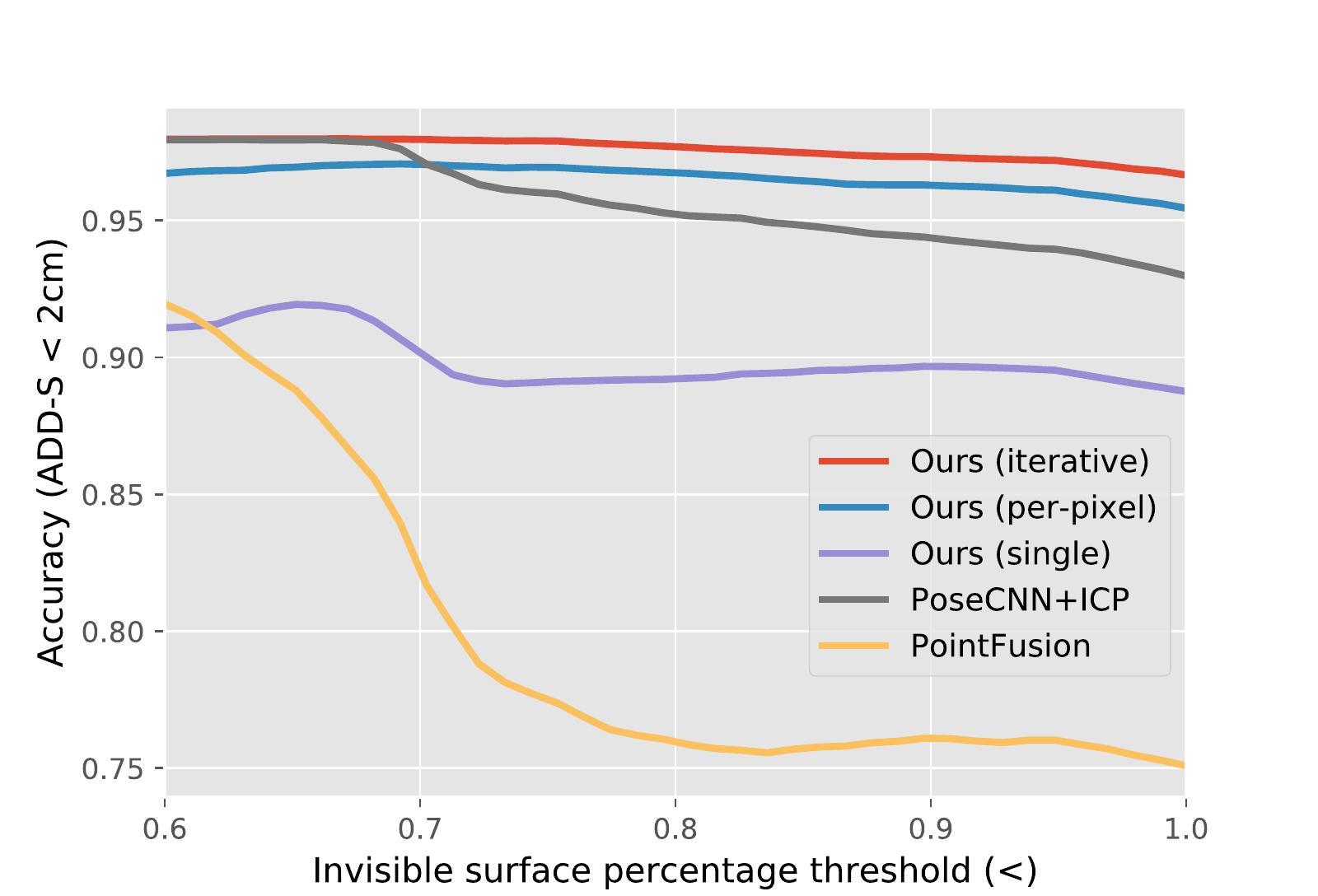}
	\caption{\textbf{Model performance under increasing levels of occlusion.} Here the levels of occlusion is estimated by calculating the invisible surface percentage of each object in the image frame. Our methods work more robustly under heavy occlusion compared to baseline methods.}
	\label{fig:exp_occlusion2}
\end{figure}

\noindent\textbf{Robustness towards occlusion} The main advantage of our dense fusion method is its robustness towards occlusions. To quantify the effect of occlusion on final performance, we calculate the visible surface ratio of each object instance (further detail available in supplementary material). Then we calculate how the accuracy (\texttt{ADD-S<2cm} percentage) changes with extent of occlusion. As shown in Fig.~\ref{fig:exp_occlusion2}, the performances of \textit{PointFusion} and PoseCNN+ICP degrade significantly as the occlusion increases. In contrast, none of our methods experiences notable performance drop. In particular, the performance of both \textit{Ours (per-pixel)} and \textit{Ours (iterative)} only decrease by $2\%$ overall.

\noindent\textbf{Time efficiency} 
We compare the time efficiency of our model with PoseCNN+ICP in Table~\ref{exp:speed}. We can see that our method is two order of magnitude faster than PoseCNN+ICP. In particular, PoseCNN+ICP spends most of time on the post processing ICP. In contrast, all of our computation component, namely segmentation (Seg), pose estimation (PE), and iterative refinement (Refine), are equally efficient, and the overall runtime is fast enough for real-time application (16 FPS, about 5 objects in each frame).

\begin{table*}
\small
\center
\caption{Quantitative evaluation of 6D pose (ADD\cite{hinterstoisser2012model}) on the LineMOD dataset. Objects with bold name are symmetric.}
\begin{tabular}{l|c|c|c|c|c|c|c} \hline
 & \multicolumn{2}{c|}{RGB} & \multicolumn{5}{c}{RGB-D} \\ \hline
 & \begin{tabular}[c]{@{}c@{}}BB8~\cite{rad2017bb8} \\ w ref.\end{tabular} & \begin{tabular}[c]{@{}c@{}}PoseCNN\\ +DeepIM\\~\cite{xiang2017posecnn,li2018deepim}\end{tabular} & \begin{tabular}[c]{@{}c@{}}Implicit\\~\cite{sundermeyer2018implicit}+ICP\end{tabular} & \begin{tabular}[c]{@{}c@{}}SSD-6D\\~\cite{kehl2017ssd}+ICP\end{tabular} & \begin{tabular}[c]{@{}c@{}}PointFusion\\~\cite{xu2017pointfusion}\end{tabular} & \begin{tabular}[c]{@{}c@{}}Ours\\ (per-pixel)\end{tabular} & \begin{tabular}[c]{@{}c@{}}Ours\\ (iterative)\end{tabular} \\ \hline
ape & 40.4 & 77.0 & 20.6 & 65 & 70.4 & 79.5 & 92.3 \\
bench vi. & 91.8 & 97.5 & 64.3 & 80 & 80.7 & 84.2 & 93.2 \\
camera & 55.7 & 93.5 & 63.2 & 78 & 60.8 & 76.5 & 94.4 \\
can & 64.1 & 96.5 & 76.1 & 86 & 61.1 & 86.6 & 93.1 \\
cat & 62.6 & 82.1 & 72.0 & 70 & 79.1 & 88.8 & 96.5 \\
driller & 74.4 & 95.0 & 41.6 & 73 & 47.3 & 77.7 & 87.0 \\
duck & 44.3 & 77.7 & 32.4 & 66 & 63.0 & 76.3 & 92.3 \\
\textbf{eggbox} & 57.8 & 97.1 & 98.6 & 100 & 99.9 & 99.9 & 99.8 \\
\textbf{glue} & 41.2 & 99.4 & 96.4 & 100 & 99.3 & 99.4 & 100.0 \\
hole p. & 67.2 & 52.8 & 49.9 & 49 & 71.8 & 79.0 & 92.1 \\
iron & 84.7 & 98.3 & 63.1 & 78 & 83.2 & 92.1 & 97.0 \\
lamp & 76.5 & 97.5 & 91.7 & 73 & 62.3 & 92.3 & 95.3 \\
phone & 54.0 & 87.7 & 71.0 & 79 & 78.8 & 88.0 & 92.8 \\ \hline
MEAN & 62.7 & 88.6 & 64.7 & 79 & 73.7 & 86.2 & 94.3 \\ \hline
\end{tabular}
\label{exp:LineMOD}
\vspace{-10pt}
\end{table*}

\noindent\textbf{Qualitative evaluation} 
Fig. \ref{exp:qualitative_ycb} visualizes some sample predictions made by PoseCNN+ICP, PointFusion, and our iterative refinement model. As we can see, PoseCNN+ICP and PointFusion fail to estimate the correct pose of the bowl in the leftmost column and the cracker box in the middle column due to heavy occlusion, whereas our method remains robust. Another challenging case is the clamp in the middle row due to poor segmentation (not shown in the figure). Our approach localizes the clamp from only the visible part of the object and effectively reduces the dependency on accurate segmentation result.

\begin{table}
\small
\centering
\caption{Runtime breakdown (second per frame on YCB-Video Dataset). Our method is approximately 200x faster than PoseCNN+ICP. Seg means Segmentation, and PE means Pose Estimation.}
\begin{tabular}{cccc|cccc}
\hline
\multicolumn{4}{c|}{PoseCNN+ICP~\cite{xiang2017posecnn}} &
\multicolumn{4}{c}{Ours}\\
 \multicolumn{1}{c}{Seg}        & \multicolumn{1}{c}{PE}        & \multicolumn{1}{c}{ICP} & ALL & \multicolumn{1}{c}{Seg}   & \multicolumn{1}{c}{PE}    & \multicolumn{1}{c}{Refine} & ALL       \\ \hline
\multicolumn{1}{c}{0.03}       & \multicolumn{1}{c}{0.17}       &  \multicolumn{1}{c}{10.4} & 10.6 & \multicolumn{1}{c}{0.03} & \multicolumn{1}{c}{0.02} & \multicolumn{1}{c}{0.01} & 0.06\\ \hline
\end{tabular}
\label{exp:speed}
\end{table}

\subsection{Evaluation on LineMOD Dataset}

\begin{figure}
	\centering
	\includegraphics[width=.95\linewidth]{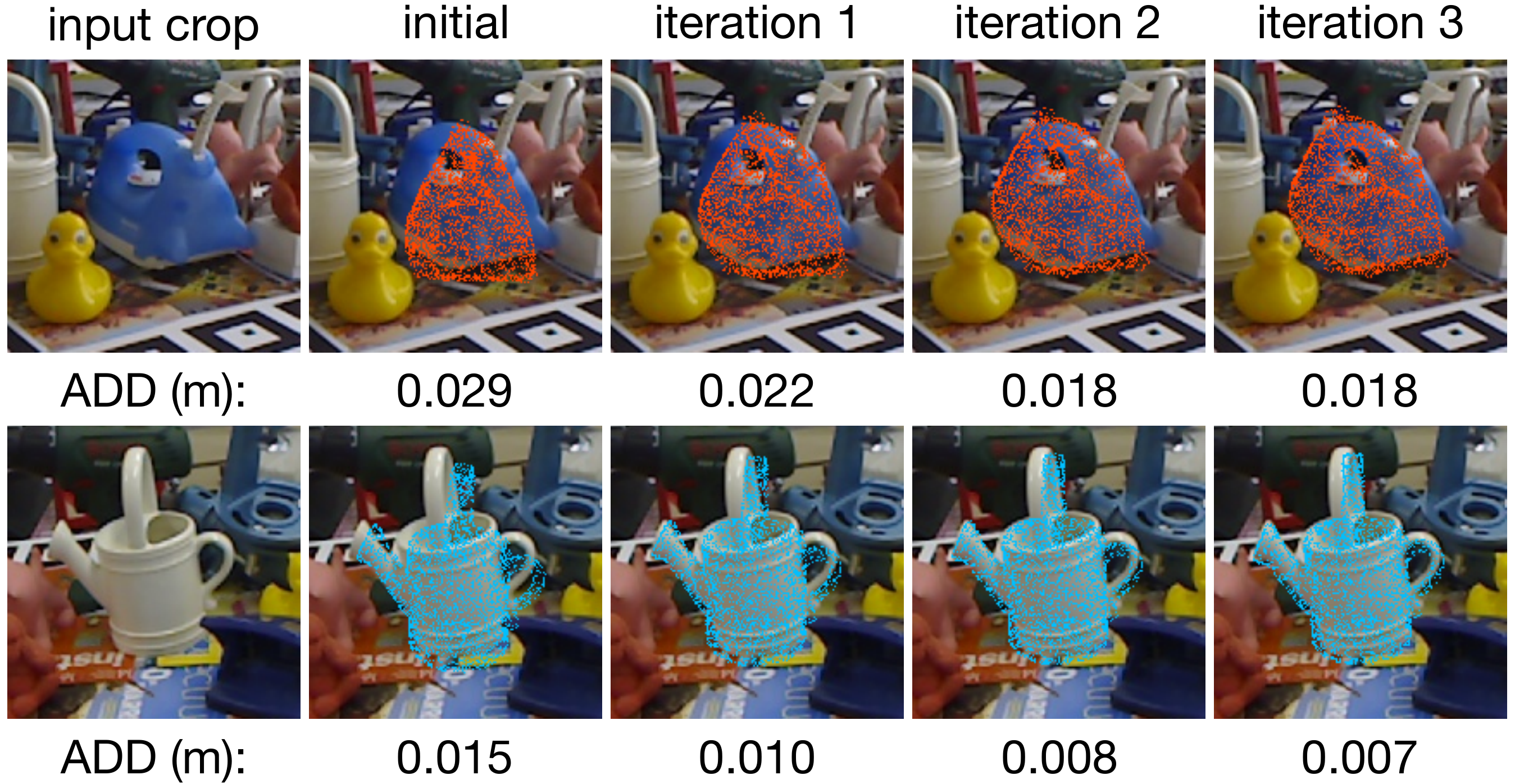}
	\caption{\textbf{Iterative refinement performance on LineMOD dataset} We visualize how our iterative refinement procedure corrects initially sub-optimal pose estimation.}
	\label{fig:linemod}
	\vspace{-10pt}
\end{figure}

Table~\ref{exp:LineMOD} compares our method with previous RGB methods with depth refinement(ICP) (results from \cite{sundermeyer2018implicit, tekin18}) on the ADD metric~\cite{hinterstoisser2012model}. Even without the iterative refinement step, our method can outperform 7\% over the state-of-the-art depth refinement method. After processing the iterative refinement approach, the final result has another 8\% improvement, which proves that our learning-based depth method is superior to the sophisticated application of ICP in both accuracy and efficiency. We visualize the estimated 6D pose after each refinement iteration in Fig.\ref{fig:linemod}, where our pose estimation improves by an average of 0.8 cm (ADD) after $2$ refinement iterations. The results of some other color-only methods are also listed in Table~\ref{exp:LineMOD} for reference.

\vspace{-5pt}
\subsection{Robotic Grasping Experiment}

In our last experiment, we evaluate whether the poses estimated by our approach are accurate enough to enable robot grasping and manipulation. As shown in Fig.~\ref{fig:pull}, we place 5 YCB objects on a table and command the robot to grasp them using the estimated pose. We follow a similar procedure to~\citet{tremblay2018deep}: we place the five objects in four different random locations on the table, at three random orientations, including configurations with partial occlusions. Since the order of picking the objects is not optimized, we do not allow configurations where objects lay on top of each other. The robot attempts 12 grasps on each object, 60 attempts in total. The robot uses the estimated object orientation to compute an alignment of the gripper's fingers to the object narrower dimension. 

The robot succeeds on 73\% of the grasps using our proposed approach to estimate the pose of the objects. The most difficult object to grasp is the banana (7 out of 12 successful attempts). One possible reason is that our banana model is not exactly the same as in the dataset -- ours is plain yellow. This characteristic hinders the estimation, especially of the orientation, and leads to some failed grasp attempts along the longer axis of the object. In spite of this less accurate case, our results indicate that our approach is robust enough to be deployed in real-world robotic tasks without explicit domain adaptation, even with a different RGB-D sensor and in a different background than the ones in the training data.

\section{Conclusion}

We presented a novel approach to estimating 6D poses of known objects from RGB-D images. Our approach fuses a dense representation of features that include color and depth information based on the confidence of their predictions. With this dense fusion approach, our method outperforms previous approaches in several datasets, and is significantly more robust against occlusions. Additionally, we demonstrated that a robot can use our proposed approach to grasp and manipulate objects. 

\section*{Acknowledgement}
This work has been partially supported by JD.com American Technologies Corporation (``JD") under the SAIL-JD AI Research Initiative and by an ONR MURI award (1186514-1-TBCJE). This article solely reflects the opinions and conclusions of its authors and not JD or any entity associated with JD.com.

\printbibliography

\newpage

\section{Supplementary Materials}

\subsection{Invisible surface percentage calculation}
The invisible surface percentage is a measurement that quantifies how occluded an object is given the camera viewpoint. The measurement is used in Sec.\ref{sec:eval_ycb} of the main manuscript. Following are the details of how to compute the invisible surface percentage.

First, we transform the ground truth model of an object to its target pose. Then, the 3D points on the surface of the model are sampled and projected back to a 2D image plane as depth pixels according to the camera intrinsic parameters. The projected depth pixels should be close to the depth measured by a depth sensor if there is \emph{no occlusion}. In other words, if the distance between the measured depth of a pixel and the model-projected depth is larger than a margin, we consider the pixel as being occluded and thus invisible. Concretely, suppose a projected depth pixel $p$ has depth value $d(p)$, and the measured depth of $p$ is $\hat{d}(p)$. $p$ is considered \emph{invisible} if $|d(p) - \hat{d}(p)| > h$. The margin $h$ is set to be 20mm in the experiment.  The invisible surface percentage is thus the percentage of the points that are \emph{invisible} out of all sampled points on the object model surface. Since around half of the points on an object model are always invisible due to self-occlusion, Fig.\ref{fig:exp_occlusion2} in the main manuscript shows results starting from $60$ invisible surface percentage.

\subsection{Details of the robotic grasping experiment}
The robot used in the experiment is a Toyota HSR (Human Support Robot). The robot is equipped with an Asus Xtion RGB-D sensor, a holonomic mobile base, and a two-finger gripper. We deployed our pose estimation model trained on YCB-Video dataset without finetuning. Note that our camera (Asus Xtion) is different from the one used to capture the YCB-Video dataset (Kinect-v2). Our experiment shows that our model is able to tolerate the difference in camera and perform accurate pose estimation. The evaluation includes five YCB objects: \texttt{005\_tomato\_soup\_can}, \texttt{006\_mustard\_bottle}, \texttt{007\_tuna\_fish\_can}, \texttt{011\_banana}, and \texttt{021\_bleach\_cleanser}.

\subsection{Additional iterative refinement examples}
\begin{figure}[h]
	\centering
	\includegraphics[width=0.95\linewidth]{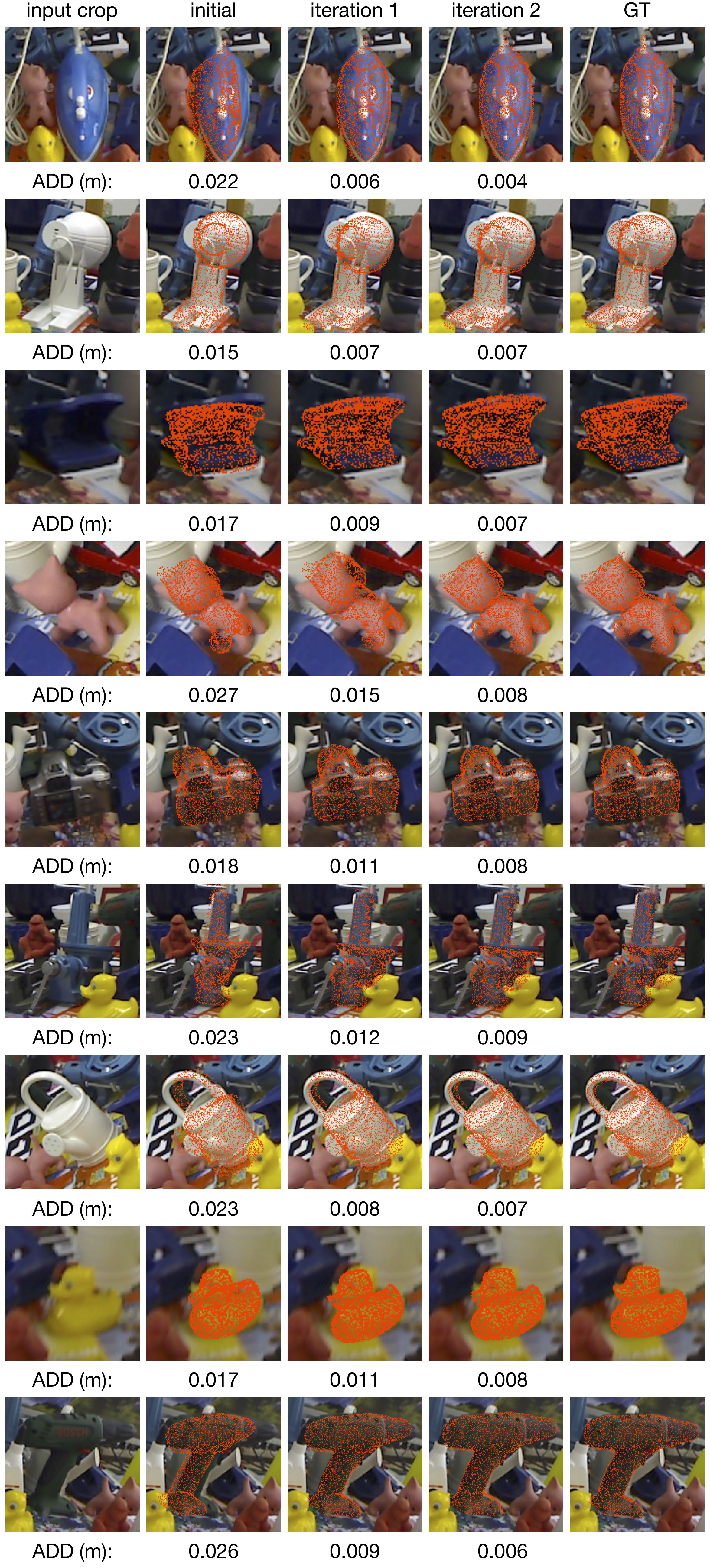}
	\caption{\textbf{Iterative refinement performance on LineMOD dataset} The initial estimation is outputted by Ours (per-pixel). We first transform the object model with the estimated pose and ground truth pose into the 3D space. The ADD distance is the average distance between each corresponding point pair on the two transformed model point clouds. Here we show our iterative refinement performance in more situations includes blurring and low light conditions, where we can see clear improvement on accuracy by using our neural network based iterative refinement method.}
	\label{exp:supp_iterative}
\end{figure}

See Fig.~\ref{exp:supp_iterative}.

\end{document}